\pdfoutput=1
\documentclass[11pt]{article}

\usepackage[preprint]{acl}
\usepackage{times}
\usepackage{latexsym}

\usepackage[T1]{fontenc}
\usepackage[utf8]{inputenc}

\usepackage{microtype}

\usepackage{inconsolata}

\usepackage{graphicx}
\usepackage{booktabs}
\usepackage{multirow}
\usepackage{url}
\usepackage{xcolor}

\usepackage{makecell} % 单元格换行
\usepackage{tabularx} % 表格自动调整列宽
\usepackage{amssymb}
\newcommand{\cmark}{\textcolor{green!70!black}{\checkmark}}  % 深绿色对勾
\newcommand{\xmark}{\textcolor{red!50!black}{\texttimes}}    % 深红色叉号
\title{DocPuzzle: A Process-Aware Benchmark for Evaluating Realistic Long-Context Reasoning Capabilities}

\author{
 \textbf{Tianyi Zhuang\textsuperscript{1}},
 \textbf{Chuqiao Kuang\textsuperscript{1}},
 \textbf{Xiaoguang Li\textsuperscript{1}},
\\
 \textbf{Yihua Teng\textsuperscript{2}},
 \textbf{Jihao Wu\textsuperscript{2}},
 \textbf{Yasheng Wang\textsuperscript{1}},
 \textbf{Lifeng Shang\textsuperscript{1}}
\\
\\
 \textsuperscript{1}Huawei Noah’s Ark Lab
 \\
 \textsuperscript{2}Huawei Inc.
\\
 \small{
   {\{zhuangtianyi, kuangchuqiao, lixiaoguang11, tengyihua, wujihao, wangyasheng, Shang.Lifeng\}@huawei.com}
 }
}

\begin{document}
\maketitle
\begin{abstract}
We present DocPuzzle, a rigorously constructed benchmark for evaluating long-context reasoning capabilities in large language models (LLMs).
This benchmark comprises 100 expert-level QA problems requiring multi-step reasoning over long real-world documents.
To ensure the task quality and complexity, we implement a human-AI collaborative annotation-validation pipeline.
DocPuzzle introduces an innovative evaluation framework that mitigates guessing bias through checklist-guided process analysis, establishing new standards for assessing reasoning capacities in LLMs.
Our evaluation results show that: 1)~Advanced slow-thinking reasoning models like o1-preview~(69.7\%) and DeepSeek-R1~(66.3\%) significantly outperform best general instruct models like Claude 3.5 Sonnet~(57.7\%); 2)~Distilled reasoning models like DeepSeek-R1-Distill-Qwen-32B~(41.3\%) falls far behind the teacher model, suggesting challenges to maintain the generalization of reasoning capabilities relying solely on distillation.
\end{abstract}

\section{Introduction}
\begin{table}[]
    \centering
    \begin{tabular}{p{7.5cm}}
    \toprule
    \textbf{Document:}
    \small ... In 2020, the wafer manufacturing materials market was \$34.9 billion. The cost of polishing materials accounted for 7\%. Over the past 20 years, the global semiconductor materials market has surged between 2001 and 2020, with a relatively even distribution among the major global economies. .... \\
    \textbf{Question:} \small Given that in 2020, the proportion of wafer manufacturing materials in the overall scale of semiconductor materials increased by 5\% … Can we
calculate the proportion of the polishing materials market in the semiconductor materials market in City X in 2019? ...\\
   \textbf{Reasoning Chain:} 
   \small \textcolor{red}{[Step1]~}\textcolor{cyan}{In 2020, the wafer manufacturing materials market was \$34.9 billion. The cost of polishing materials accounted for 7\%.}
   The market size was \colorbox{cyan}{34.9} × \colorbox{cyan}{0.07} = \$2.443 billion.
   \textcolor{red}{[Step2]~}\textcolor{cyan}{In 2020, the market size of wafer manufacturing materials ..., an increase of 6.5\% compared to 2019. The cost of polishing materials accounted for 4.41\% of the total cost of semiconductor materials.}
   The overall semiconductor materials market was \colorbox{yellow}{2.443} ÷ \colorbox{cyan}{0.0441} = \$55.3968 billion.
\textcolor{red}{[Step3]~}\textcolor{green}{The polishing materials market remained unchanged}, the polishing materials market in 2019 was also \$2.443 billion, and it was all produced in City X.
\textcolor{red}{[Step4]~}The wafer manufacturing market in 2019 was \colorbox{cyan}{34.9} ÷ \colorbox{cyan}{1.065} = \$32.77 billion.
\textcolor{red}{[Step5]~}The proportion of wafer manufacturing in semiconductor materials in 2020 was \colorbox{cyan}{34.9} ÷ \colorbox{yellow}{55.3968} = 63\%, \textcolor{red}{[Step6]~}so this proportion was \colorbox{yellow}{63\%} - \colorbox{green}{5\%} = 58\% in 2019.
\textcolor{red}{[Step7]~}\textcolor{cyan}{City X had the largest share, reaching 22\% in 2019.}
\textcolor{red}{[Step8]~}The semiconductor market was \colorbox{yellow}{32.77} ÷ \colorbox{yellow}{0.58} = \$56.5 billion, of which City X‘s semiconductor market is \colorbox{yellow}{56.5} * \colorbox{cyan}{22\%} = \$12.43 billion.
\textcolor{red}{[Step9]~}In City X in 2019, the polishing materials market was \$2.443 billion, and the semiconductor market was \$12.43 billion, with a ratio of \colorbox{yellow}{2.443} / \colorbox{yellow}{12.43} = 19.65\%.
\\
\textbf{Answer:} \small The ratio is 19.65\%.\\
\textbf{Checklist:} \small  Calculation error is allowed, so an answer between 18\% and 21\% is considered correct. 
The response must include that the polishing market in 2019 was (about) 2.4 billion, and the semiconductor market in 2019 was (about) 56.5 billion. \\
 \bottomrule
    \end{tabular}
    \caption{An example of question, answer, checklist, reasoning chain pairs, translated to English. The reasoning chain is utilized for human validation.
    Related paragraphs in documents are in \textcolor{cyan}{cyan}, and related information in question is in \textcolor{green}{green}.
    Statistics highlighted in \colorbox{yellow}{yellow} are inferred from existing steps, those in \colorbox{cyan}{cyan} are retrieved from documents, and those in \colorbox{green}{green} are retrieved from the question.}
    \label{tab:sample}
\end{table}

The rapid evolution of large language models (LLMs) \cite{openai_o1preview, claude35, deepseekai2025deepseekr1incentivizingreasoningcapability} has demonstrated unprecedented capabilities in long-context processing and complex reasoning. 
These advancements extend the boundaries of machine intelligence and reignite discussions about achieving Artificial General Intelligence (AGI). 
However, current technical reports of these LLMs predominantly focus on structured mathematical problem-solving \cite{MATH,lightman2023lets,AIME} and coding tasks \cite{LiveCodeBench}, creating a significant disconnect between benchmark performance and real-world reasoning requirements. 
A critical gap in assessing models' capacity for extended logical chaining with implicit connections remains understudied.

We focus on the design of long-context reasoning benchmarks based on the following issues in real-world scenarios. 
The parametric knowledge encapsulated during training inevitably suffers from temporal limitations. 
However, practical scenarios frequently demand analysis of emerging concepts or private documents that never existed in the training corpus. 
Long-context reasoning capability enables LLMs to assimilate such through contextual understanding while performing coherent reasoning.
Moreover, this capability exhibits essential domain adaptability as documents can come from various domains.

Existing long-context benchmarks mostly emphasize retrieval capabilities over genuine reasoning \cite{BAMBOO, RULER}. 
Typical evaluation paradigms exhibit three fundamental limitations: 1) Oversimplified reasoning where solutions require single-step evidence retrieval \cite{Vodrahalli2024MichelangeloLC, DOCBENCH} or shallow logical operations \cite{Vodrahalli2024MichelangeloLC, DOCBENCH, INFINITEBENCH, HELMET, DocFinQA}; 
2) Format-driven evaluation that prioritizes evaluation convenience, restricting response formats to multiple-choice or true-false questions, introducing significant guessing biases\cite{LongBench2, NoCha}; 
and 3) Domain monotonicity focusing on narrow verticals like literature analysis \cite{NoCha, DetectiveQA} while neglecting the generalization ability. 
These limitations undermine the discriminative power of existing evaluations, particularly for cutting-edge models approaching human-level performance on traditional benchmarks.

To address these gaps, we present DocPuzzle, a Chinese long-context reasoning benchmark consisting of 100 QA tasks over long real-world documents.
DocPuzzle is characterized by the following features: 
(1)~\textbf{Multiple realistic domains}. Documents in DocPuzzle are sourced from five diverse domains including academic papers, financial reports, etc. 
(2)~\textbf{Challenging reasoning}. Questions in DocPuzzle necessitate multi-step reasoning operations including arithmetic reasoning, temporal reasoning, etc. We also employ human-AI collaborative validation to ensure quality and challenge level for state-of-the-art LLMs.
(3)~\textbf{Process-aware evaluation}. A sample in DocPuzzle consists of a document, question, answer, and checklist. The checklist will be used during evaluation to check if the reasoning process is correct, thus mitigating the guessing bias of LLMs. An example is shown in Table \ref{tab:sample}).

Our contributions can be summarized as follows:
\begin{itemize}
    \item  \textbf{DocPuzzle Benchmark}:  A human-annotated Chinese dataset featuring 100 multi-domain cases with verification mechanisms. 
    % \footnote{The dataset will be publicly available soon}
    Each case integrates knowledge grounding, evidence chaining, and reasoning requirements.
    \item \textbf{Process-Aware Evaluation Framework}: A novel checklist-based assessment system that decouples reasoning validity from final answer correctness. 
    To the best of our knowledge, we are the first to offer robust, systematic model-based evaluations for the reasoning process without constraining response formats.
    \item \textbf{Competency Gap Analysis}: Systematic profiling of reasoning capabilities of LLMs.
    Our analysis reveals the reasoning capability of the models widely used in the research community.
\end{itemize}

\section{Related Works}
\setlength{\tabcolsep}{3pt}
\begin{table*}[]
    \centering
\begin{tabular}{lcccccc}
\toprule  
 & \centering \makecell{\small Long-Context \\ \small Based} & \makecell{\small Realistic Tasks\\ \small Included}  & \makecell{\small Challenging\\ \small Reasoning} & \makecell{\small Free-form\\ \small Answer} & \makecell{\small Controversy\\ \small Review } & \makecell{\small Process-aware\\\small Evaluation} \\
\midrule
AIME\textsuperscript{\ref{footnote:AIME}}&  \xmark & -- & \cmark & \cmark & -- & \xmark\\
HLE\cite{HLE} & \xmark & -- & \cmark & \cmark & \cmark & \xmark\\
Ruler\cite{RULER} & \cmark & \xmark & \xmark & \cmark & \xmark & \xmark\\
% LongBench & \cmark & \cmark & \xmark & \cmark & \xmark & \xmark\\
InfiniteBench\cite{INFINITEBENCH} & \cmark & \cmark & \xmark & \cmark & \xmark & \xmark\\
HELMET\cite{HELMET} & \cmark & \cmark & \xmark & \cmark & \xmark & \xmark\\
BABILong\cite{BABILong} & \cmark & \xmark & \xmark & \cmark & \xmark & \xmark\\
LongBench v2\cite{LongBench2} & \cmark & \cmark & \cmark & \xmark & \cmark & \xmark\\
\midrule
DocPuzzle(Ours)& \cmark & \cmark & \cmark & \cmark & \cmark & \cmark\\
\bottomrule
\end{tabular}
    \caption{Comparison of long-context benchmarks and reasoning benchmarks.  The columns are defined as follows:
(1)~Long-Context Based: Includes a lengthy context to answer upon; 
(2)~Realistic Tasks: Simulates real-world applications (e.g., document QA), excluding synthetic tasks; 
(3)~Challenging Reasoning: Requires deliberate multi-step reasoning that tests the slow-thinking capabilities of LLMs; 
(4)~Free-form Answer: Includes tasks whose answers are not predefined choices; (5)~Controversy Review: Incorporate independent reviewers in the annotation pipeline to identify controversial answers; (6)~Process-aware Evaluation: Assesses intermediate steps rather than solely final outputs to prevent random guesses.  }
    \label{tab:related_works}
\end{table*}

\subsection{Context-free Reasoning Benchmarks}
Recently, with the release of OpenAI’s o1 \cite{o1report}, there has been growing attention on slow thinking and long-form Chain of Thought (CoT) as a means of improving reasoning capabilities \cite{deepseekai2025deepseekr1incentivizingreasoningcapability,qwq-32b-preview,Qin2024O1RJ}. These efforts primarily focus on domains that demand rigorous logical reasoning:
(1) Competitive math problems: AIME\footnote{https://huggingface.co/datasets/Maxwell-Jia/AIME\_2024\label{footnote:AIME}}, CNMO\footnote{https://www.cms.org.cn/Home/comp/comp/cid/12.html} and MATH\cite{MATH}; (2) Coding challenges: SWE-Bench \cite{SWE-bench}, LiveCodeBench \cite{LiveCodeBench} and Codeforces\footnote{https://codeforces.com}; (3) Scientific academic problems: GPQA \cite{GPQA} and Humanity's Last Exam  \cite{HLE}.

While these tasks address reasoning difficulties at the forefront of human intelligence, 
we do not regard these domain-specific reasoning benchmarks as adequate measures of the generalization ability of LLMs. With the intuition that diverse domains can naturally emerge from a wide range of documents, we pay attention to long-context documents to present problems that span an extensive range of fields.

\subsection{Long-context Reasoning Benchmarks}
\paragraph{Elementary Long-context Benchmarks} In the early stages of the development of long-context LLMs, synthetic tasks \cite{NIAH, RULER} dominated the evaluation landscape as a basic measure of long-context understanding. Recognizing the potential divergence between the results of synthetic tasks and real-world applications, LongBench \cite{LongBench} and InfiniteBench \cite{INFINITEBENCH} incorporate realistic tasks, such as document-based question answering and summarization. To achieve better discrimination, HELMET \cite{HELMET} aggregates contemporary long-context benchmarks. The aforementioned studies focus primarily on evaluating the fundamental capacity for long-context comprehension. Reasoning questions are scarcely incorporated, and when present, they tend to be rudimentary, requiring only one or two steps.

Although several studies explicitly emphasize evaluating long-context reasoning, their ability to discriminate is often limited. BABILong \cite{BABILong} assesses reasoning abilities through synthetically generated extended contexts. DocBench \cite{DOCBENCH} and Loong \cite{Loong} incorporate reasoning tasks based on single or multiple documents. These benchmarks face two key limitations: (1) compromised validity due to semantically inconsistent synthetic contexts, or (2) excessively simplified questions that fail to pose genuine challenges to LLMs proficient in slow and deliberate reasoning.

\paragraph{Difficult Long-context Reasoning Benchmarks} Recent works have introduced more challenging tasks that increase the complexity of evidence retrieval or require longer chains of logical reasoning. Nocha \cite{NoCha}, RuleArena \cite{RuleArena}, and DetectiveQA \cite{DetectiveQA} include difficult reasoning tasks that focus on specific domains such as fantasy novels, policies, and orthodox detective stories, respectively. LongBench v2 \cite{LongBench2} introduces difficult human-curated long-context reasoning tasks across various domains. However, their discriminative validity is undermined by three key factors: (1) limited domain coverage \cite{NoCha,RuleArena,DetectiveQA}; (2) reduced reasoning difficulty due to the exclusion of deliberately designed challenging puzzles and the reliance on domain-specific expertise \cite{LongBench2}; (3) the use of a unified multiple-choice question format, which allows the model to guess the correct answer randomly \cite{LongBench2}.
We compare our DocPuzzle benchmark and others in Table \ref{tab:related_works}.

\section{DocPuzzle}

\begin{figure*}
    \centering
    \includegraphics[width=\linewidth]{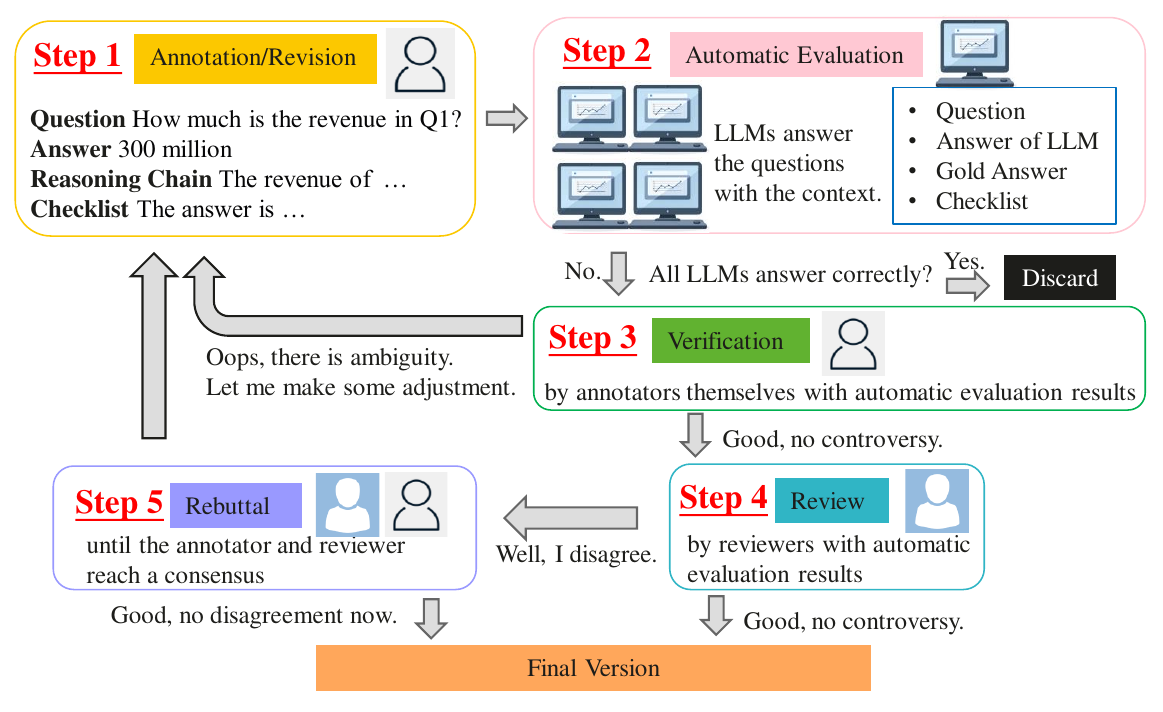}
    \caption{Overview of DocPuzzle construction pipeline. There are five stages before the samples are included in the benchmark.}
    \label{fig:annotation_pipeline}
\end{figure*}
DocPuzzle is a benchmark crafted to evaluate the ability of LLMs to handle highly intricate reasoning tasks based on long contexts. 
Figure \ref{fig:annotation_pipeline} illustrates the carefully designed data annotation process, which emphasizes ensuring the difficulty of the puzzles and incorporates a meticulous verification of answer controversy. 
Section \ref{sub:pipeline} introduces three stages in detail.

\subsection{Data Construction Pipeline}\label{sub:pipeline}

\begin{figure}
    \centering
    \includegraphics[width=0.8\linewidth]{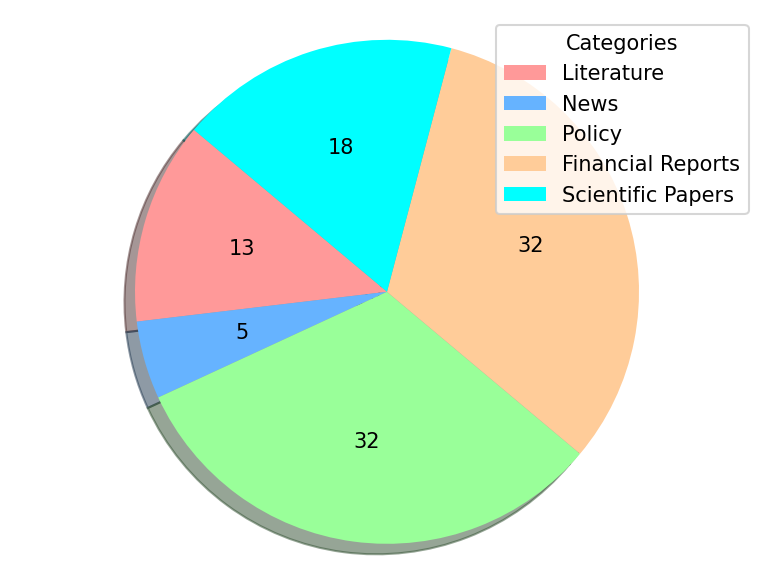}
    \caption{Distribution of document categories in DocPuzzle benchmark.}
    \label{fig:context_distribution}
\end{figure}

\paragraph{Data Collection}
We curate contexts from five domains: literature, news articles, policy documents, financial reports, and scientific papers.
The domain distribution of collected contexts is visualized in Figure \ref{fig:context_distribution}.

We recruit domain experts with at least a master's degree as annotators. To ensure sufficient reasoning depth, these annotators upload reasoning-intensive documents that they are highly familiar with. 
For policy documents and news reports requiring extended context, annotators perform article aggregation by retrieving related documents and concatenating them into coherent narratives. 
Contexts undergo semantic integrity verification through manual review, with necessary modifications to ensure answer determinism. 
The final dataset contains contexts with median and average token counts of 10,641 and 10,215 in respective \footnote{Tokenized using Qwen2.5-72B-Instruct tokenizer}.

\paragraph{Data Annotation}
Annotators follow rigorous guidelines that adhere to the following principles.
1)Context-Dependent Reasoning: 
In all cases,  the information required to answer the questions has to be inferred from the relevant content of the original documents rather than solely inferred from common sense or pre-trained knowledge patterns.
This design ensures the evaluation of models' contextual reasoning abilities rather than memorization capacity, as LLMs have been trained with numerous corpora, including reasoning-intensive ones.
2) Reasoning Complexity: Each question must involve at least two reasoning operations from the following categories: temporal reasoning, arithmetic reasoning, bridging, comparative analysis, or causal inference. 
We explicitly prohibit answerable questions through single-fragment retrieval.
Following \cite{geva2021didaristotleuselaptop}, we prioritize implicit reasoning paths without direct solution shortcuts. 
While allowing the incorporation of common sense knowledge, we intentionally include cognitive traps to assess model robustness under realistic confusion scenarios.
3)Answer Objectivity: To minimize evaluation bias, we restrict questions to objective ones (multiple-choice, true/false, direct Q\&A) with deterministic answers. 
Subjective questions requiring interpretative responses are excluded due to potential annotation variance.

\paragraph{Data Verification and Revision}
To address the natural language ambiguity in complex reasoning tasks, we implement a validation protocol in which the final answer gradually converges through multiple iterations of independent responses, debate, and revision. 
In the first stage, problems are answered using state-of-the-art LLMs (o1-preview\footnote{o1-preview-2024-09-12}, GPT-4o\footnote{gpt-4o-2024-08-06 \label{footnote:4o}}, QwQ-32B-Preview \cite{qwq-32b-preview}, moonshot-v1-128k \footnote{https://platform.moonshot.cn \label{footnote:kimi}}).
The responses are evaluated following the methodology in Section \ref{subsec:eval_method}. 
Annotators analyze cases especially when all models fail and revise samples with controversial answers and checklists.
Besides, we implement difficulty filtering. 
Samples solved by all baseline models are removed to maintain challenge levels. 

In the second stage, each sample undergoes independent human review. 
In cases of inconsistency, a debate between the reviewer and the annotator ensues to establish a rigorous, objective answer. 
Inconsistencies are typically due to missing information, subjective deductions, or ambiguity in the content of a document or problem. 
Contested items require consensus through debate. 
Samples with persistent disagreements are discarded.
The final dataset retains only samples demonstrating stable expert consensus.

\subsection{Evaluation Method}\label{subsec:eval_method} 
Existing reasoning benchmarks predominantly focus on structured domains like mathematics and coding \cite{lightman2023lets, jain2024livecodebenchholisticcontaminationfree} or utilize constrained formats \cite{LongBench2}, limiting their applicability to free-form textual reasoning.
As we focus on the capacity of reasoning, we allow imperfect textual organizing such as language mixing and redundant content.
Under this circumstance, employing a parser for the response may introduce bias.
On the other hand, even an erroneous reasoning chain may lead to a correct answer.
Thus, a model good at guessing will get an inflated score, especially for multiple-choice and true-or-false questions.
Our evaluation framework addresses these limitations.
To mitigate these biases, we take not only the final answer but also the reasoning chain into account by utilizing a checklist for each sample.
We allow a calculation error within an acceptable range in the checklist.
This accommodates acceptable calculation deviations while penalizing fundamental logical errors.
We prompt a judge model to judge whether the response is correct. 
To mitigate randomness, we design three similar evaluation prompts, and the evaluation result is decided by majority voting.
The prompt templates we use in the evaluation are shown in Appendix \ref{sec:appendix}.

\section{Experiments}
\begin{table*}[]
    \centering
    \begin{tabular}{clccc}
   \toprule  
 Model Type  & Model  & Score & Total & Accuracy(\%) \\
\hline
\multirow{3}{*}{Reasoning Model}   & o1-preview   & 69.7 & 100 & \underline{69.7} \\
 & DeepSeek-R1 & 66.3 & 100 & 66.3 \\
  & DeepSeek-R1-Distill-Qwen-32B & 39.7 & 100 & 39.7 \\
 & QwQ-32B-Preview  & 41.3 & 100 & 41.3 \\
 \hline
 \multirow{15}{*}{Instruct Model}  & GPT-4o & 41 & 99 & 41.3 \\
  &  \ \ \  $_{+\  COT\ prompt}$ & 43.7 & 100 & 43.7(+2.4) \\
   & Claude 3.5 Sonnet & 48.3 & 100 & \underline{48.3} \\
  & \ \ \  $_{+\  COT\ prompt}$ & 57.7 & 100 & \underline{57.7(+9.4)} \\
   &  DeepSeek-V3 & 41.7 & 100 & 41.7 \\
 & \ \ \ $_{+\  COT\ prompt}$ & 46 & 100 & 46(+4.3) \\
&  Qwen2.5-72B-Instruct & 39.7 & 100 & 39.7 \\
 & \ \ \ $_{+\  COT\ prompt}$ & 45 & 100 & 45(+5.3) \\
 & Qwen2.5-32B-Instruct & 32.7& 100&32.7 \\
& \ \ \ $_{+\  COT\ prompt}$ & 36.3&100&36.3(+3.6) \\
& Qwen2.5-14B-Instruct & 27 & 100 & 27\\
& \ \ \ $_{+\  COT\ prompt}$ &27.3 &100&27.3(+0.3) \\
&  Qwen2.5-7B-Instruct & 20.3 &100 &20.3 \\
& \ \ \ $_{+\  COT\ prompt}$ &17.3&100 & 17.3(-3) \\
 & moonshot-v1-128k & 28 & 97 & 28.9 \\ 
   & \ \ \ $_{+\  COT\ prompt}$ & 25 & 97 & 25.7(-3.2)\\
 \bottomrule
    \end{tabular}
    \caption{Main results of different LLMs on DocPuzzle. Best scores are \underline{underlined}}
    \label{tab:main_results}
\end{table*}

\subsection{Baselines}
We conduct comprehensive evaluations across two model categories:

\paragraph{Slow-Thinking Reasoning Models} We evaluate the following four slow-thinking reasoning models: 
o1-preview \cite{openai_o1preview}, DeepSeek-R1\cite{deepseekai2025deepseekr1incentivizingreasoningcapability}, DeepSeek-R1-Distill-Qwen-32B\cite{deepseekai2025deepseekr1incentivizingreasoningcapability}, QwQ-32B-Preview\cite{qwq-32b-preview}.
For DeepSeek-R1, the response consists of reasoning content and the response content, so we take both into account\footnote{For more details, see \url{https://api-docs.deepseek.com/guides/reasoning_model}}.
For DeepSeek-R1-Distill-Qwen-32B, we set temperature to 0.3, top\_p to 0.2, and top\_k to 0.

\paragraph{Instruct Models} We evaluate the following instruct models: GPT-4o\cite{openai_gpt4o}, Claude 3.5 Sonnet \cite{claude35}, DeepSeek-V3\cite{deepseekai2024deepseekv3technicalreport}, Qwen2.5-72B-Instruct\cite{qwen2.5} and moonshot-v1-128k \textsuperscript{\ref{footnote:kimi}}.
We do not evaluate the performance of the LLaMA series, as questions in our benchmark are mostly Chinese.
For instruct models, we also evaluate the performance when employing the powerful zero-shot COT prompt, "Let's think step-by-step" \cite{kojima2023largelanguagemodelszeroshot}.
\paragraph{Setups} We use GPT-4o\textsuperscript{\ref{footnote:4o}} as the judge model.
As demonstrated in Section \ref{subsec:eval_method}, we utilize majority voting across three prompt variants.
Furthermore, to mitigate biases, we run the evaluation scripts three times, and the score is the average of the three results.

\subsection{Main Results}
Due to factors like risk control, not all requests may get valid responses.
We present the main results in Table \ref{tab:main_results}.
\paragraph{Reasoning Capability Ranking}
Slow-thinking reasoning models like o1-preview and DeepSeek-R1 significantly outperform other models on the benchmark.
DocPuzzle shows consistent rankings for the advancing models with other reasoning benchmarks, demonstrating the reasoning nature of DocPuzzle.
While most instruction models underperform specialized slow-thinking reasoning models, Claude 3.5 Sonnet and DeepSeek-V3 demonstrate exceptional reasoning capability, surpassing the Qwen-based models.

\paragraph{Scaling Law Manifestation} 
Model capability exhibits a strong correlation with reasoning ability and domain knowledge. 
Qwen2.5 series shows consistent accuracy gains from 7B (20.3\%) to 72B (39.7\%) variants.

\paragraph{Contrastive Prompt Effects}
While zero-shot CoT prompting \cite{kojima2023largelanguagemodelszeroshot} generally improves performance, effectiveness diminishes for models with limited reasoning performance. 
CoT prompting yields maximum gains for Claude 3.5 Sonnet and Qwen2.5-72B-Instruct.
Performance degradation is observed in Qwen2.5-7B-Instruct and moonshot-v1.
CoT effectiveness emerges when instruct models achieve a score over 32.7\%, indicating minimum model capacity requirements for reasoning path utilization.

\section{Analysis}
\subsection{Response Length}
\begin{figure}
    \centering
    \includegraphics[width=\linewidth]{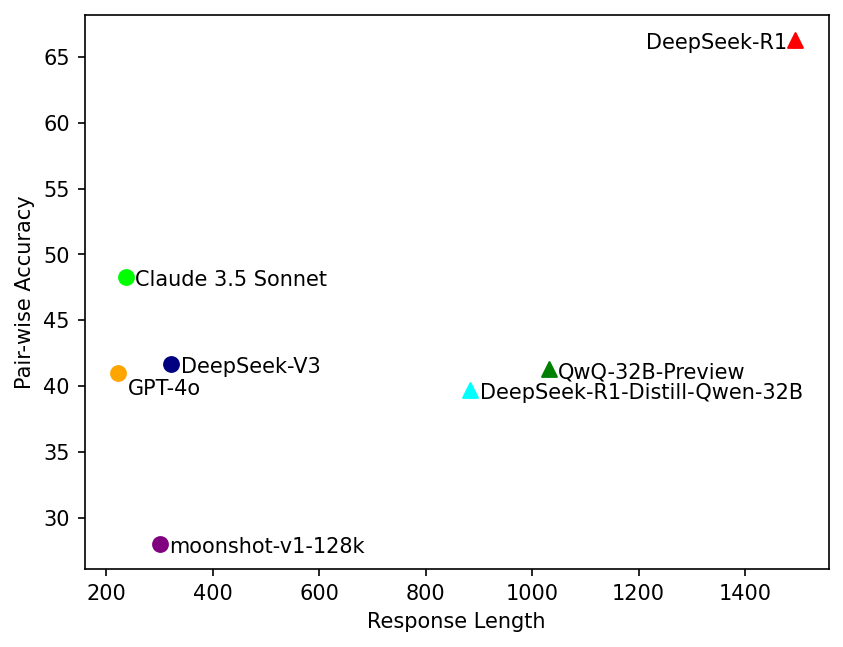}
    \caption{Response length vs. accuracy relationship across model types.}
    \label{fig:len_performance}
\end{figure}

We conduct an analysis of response length\footnote{Using Qwen2.5-72B-Instruct tokenizer}  and its correlation with accuracy metrics as illustrated in Figure \ref{fig:len_performance}.
The o1-preview model is excluded from this analysis due to the unavailability of its reasoning chain.
A significant disparity emerges between the average lengths of reasoning models and instruct models.
Notably, while distilled models exhibit substantially increased response lengths that do incorporate reasoning chains, a persistent gap remains between distilled models and their teacher counterparts in terms of length-accuracy optimization.

\subsection{Is the reasoning ability of distilled models generalizable?}
\begin{figure}
    \centering
\includegraphics[width=\linewidth]{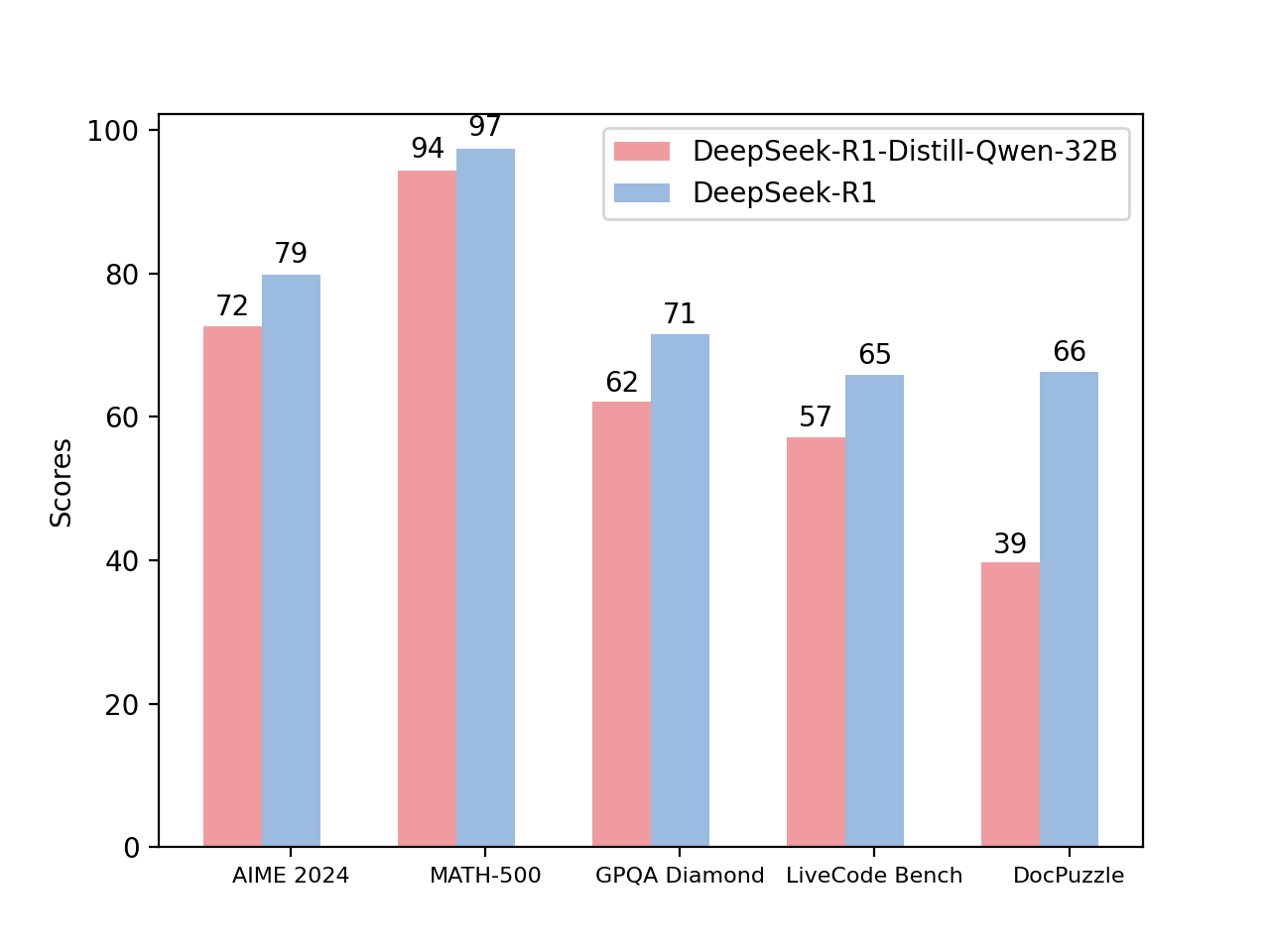}
    \caption{The performance of DeepSeek-R1 and DeepSeek-R1-Distill-Qwen-32B on different benchmarks.}
    \label{fig:distill-teacher}
\end{figure}

Though DeepSeek-R1-Distill-Qwen-32B achieves comparable scores on math and code datasets\cite{deepseekai2025deepseekr1incentivizingreasoningcapability}, its efficacy substantially degrades on the DocPuzzle benchmark(see Figure \ref{fig:distill-teacher}). 
Furthermore, as evidenced in Table \ref{tab:main_results}, the gap between DeepSeek-R1-Distill-Qwen-32B and Qwen2.5-32B-Instruct with a zero-shot COT prompt remains statistically insignificant.
These findings suggest that reasoning capabilities acquired through supervised finetuning on distilled datasets produced by heterogeneous models exhibit limited generalization to complex, long-context reasoning scenarios.
This conclusion aligns with \citet{chu2025sft}, who demonstrate that models trained by supervised fine-tuning tend to memorize surface patterns in training data while struggling with out-of-distribution generalization. 
The empirical evidence challenges the universal applicability of knowledge distillation techniques and calls for other techniques,  such as reinforcement learning and self-improvement, to improve reasoning generalization.

\subsection{The Potential of Exploration}
\begin{figure}
    \centering
    \includegraphics[width=\linewidth]{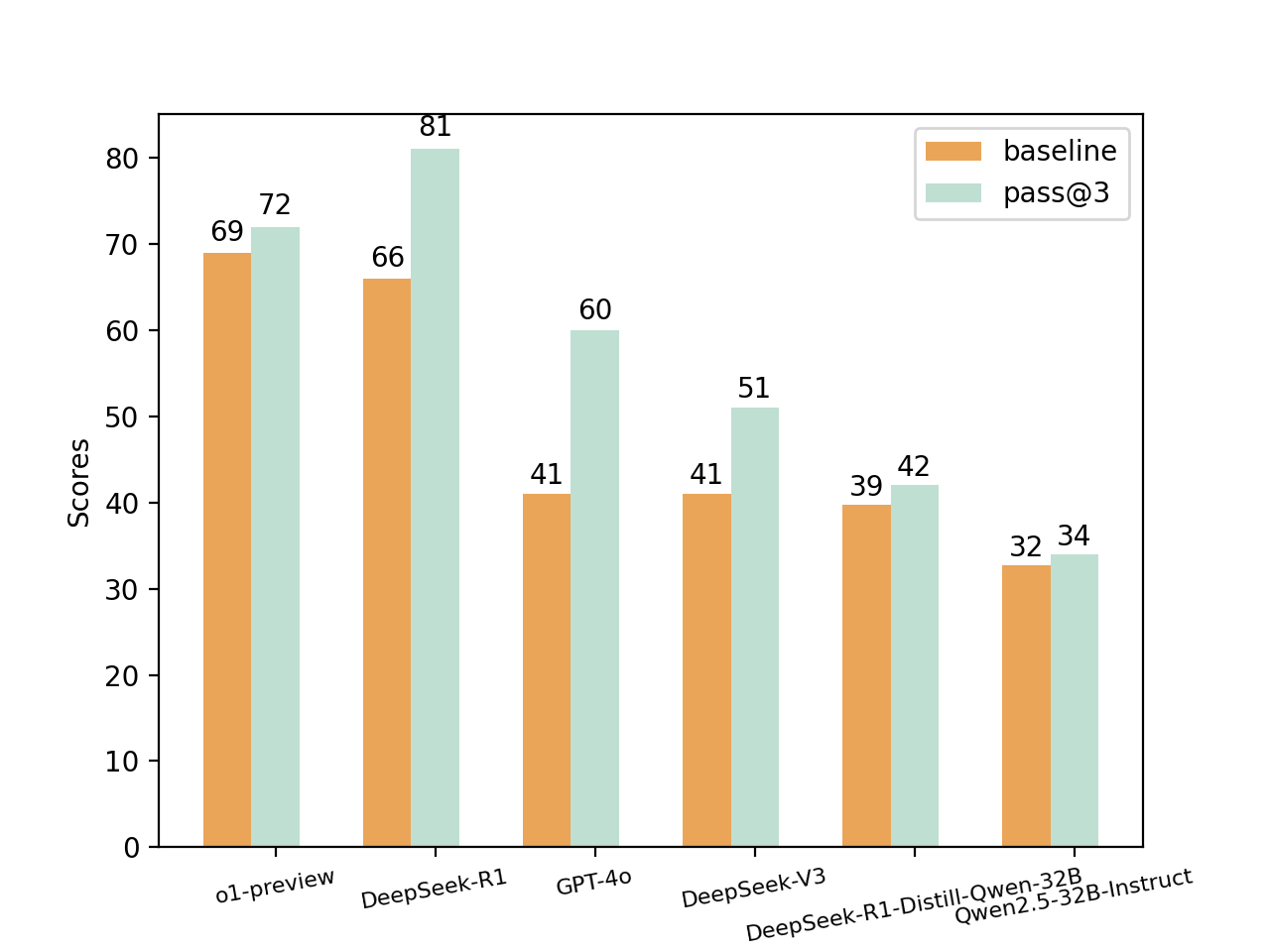}
    \caption{Measuring exploration potential via pass@3 metric.}
    \label{fig:pass_at_3}
\end{figure}

Given the growing prominence of reinforcement learning (RL) in language model training, we quantify exploration capacity through the pass@3 score.
Figure \ref{fig:pass_at_3} illustrates that DeepSeek-R1 achieves superior exploration capability.
Our analysis reveals no significant correlation between LLM exploration potential (the difference between pass@3 and baseline scores) and pairwise accuracy scores.
We extend this investigation to DeepSeek-R1-Distill-Qwen-32B and Qwen2.5-32B-Instruct, derived from the Qwen2.5-32B base model.
Under controlled experimental conditions (temperature = 0.3, top\_p = 0.2, top\_k = 0), both models exhibit constrained exploration capacities as shown in Figure \ref{fig:pass_at_3}.

\begin{table*}[]
\centering
\begin{tabular}{cp{3cm}p{5.6cm}p{4.5cm}}
\toprule
Reasoning Type & Question &o1-preview Answer & DeepSeek-R1 Answer \\
\midrule
Arithmetic & \small The spot table has an error in one of the soybean oil prices, and the basis table has an error in one of the rapeseed oil prices. Please identify and correct these errors. Then, ...
& 
\small ...
Error in the spot price table (soybean oil):
The spot price of Grade 1 soybean oil in Tianjin was listed as 6,110.00, which is significantly lower than in other regions and appears to be an input error. \textcolor{red}{Considering that the prices in other regions range from 6,650.00 to 6,790.00, and the price change in Tianjin is 30.00, the reasonable price should be 6,710.00 (consistent with Rizhao). Therefore, the spot price of Grade 1 soybean oil in Tianjin should be corrected from 6,110.00 to 6,710.00.} ... &
\small ... Spot Price Table Error:
The spot price of Grade 1 soybean oil in Jiangsu was incorrectly recorded as 9,100.00(-50.00). \textcolor{red}{Based on the context, Jiangsu should be the region for Grade 4 rapeseed oil, and the correct price for soybean oil should be similar to other regions (such as Tianjin). Correction:
6,100.00(-50.00) (assuming the previous price was 6,150.00, with a decrease of 50 yuan in this period).}\\
\hline
Common Sense & \small Which of the following institutions are definitely unable to register as Shanghai vocational skills training institutions?
A. ...
B. The training site has a usable area of 140 square meters.
C. ...
D. ... &
\small 
...
Option B:
The training site is in a unit with a usable area of 140 square meters.
According to the 14th section "Training Sites and Temporary Activity Venues" of the document, the gross floor area of the training site should be no less than 150 square meters. \textcolor{red}{Therefore, the training site's area of only 140 square meters does not meet the minimum area requirement.} Consequently, Option B cannot be registered.
Option C:
... &
\small
...
Option B: "The training site has a usable area of 140 square meters." According to Article 14 regarding the training site, the gross floor area of the training site must be no less than 150 square meters. \textcolor{red}{The 140 square meters given here is clearly insufficient, so Option B does not comply and cannot be registered.}
Option C: ...
\\
\bottomrule
\end{tabular}
\caption{Error Case Analysis: the erroneous reasoning chain is in \textcolor{red}{red}.}
\label{tab:error}
\end{table*}

\subsection{Typical Error Analysis}
To investigate the limitations of the reasoning capacity of state-of-the-art LLMs, we conduct an error analysis for o1-preview and DeepSeek-R1 in Table \ref{tab:error}.
\paragraph{Arithmetic Reasoning Deficits}
The answer can be inferred by comparing two tables of futures prices and spot prices: (1)~Spot Price = Basis + Futures Price; 
(2)~As the equation does not hold where 
Zhangjiagang spot price = 6650, basis = 406, futures price = 6344,
and Hubei rapeseed oil spot price = 9350, basis = 661, futures price = 8449.
(3)~According to the question, there is an error in one of the soy-bean oil prices in the spot table, and there is an error in one of the rapeseed oil prices in the basis table.
Hence, the error in the spot table should be corrected as Zhangjiagang spot price =406 + 6344 = 6750.
The error in the basis table should be corrected as Hubei rapeseed oil basis = 9350 - 8449 = 901.  
Both models fail to extract the price relationship formula from tabular data, instead generating plausible but incorrect responses through heuristic estimation.
\paragraph{Common Sense Reasoning Limitations}
It is common knowledge that the gross floor area is larger than the usable area.
However, o1-preview and DeepSeek-R1 fail to recognize this difference, leading to incorrect judgments.
The lack of certain common sense knowledge prevents LLMs from engaging in further implicit reasoning.

\section{Conclusions}
We introduce DocPuzzle, a challenging yet realistic long-context reasoning benchmark. By a carefully designed construction pipeline, DocPuzzle ensures both reliability and difficulty and shows consistent rankings for advanced LLM models.
Instead of only considering the final answer, a process-aware evaluation method is also proposed to better evaluate the reasoning process of LLMs.
Our study through DocPuzzle reveals the strong long-context reasoning ability of slow-thinking models.
Another insight is that Knowledge distillation may be insufficient for transferring complex reasoning patterns.

\section*{Ethics Statement}
DocPuzzle seeks to establish an instructive benchmark for advancing research in complex reasoning with long context.
Certain data incorporated within DocPuzzle are sourced from open-source datasets and publicly available data on the Internet.
The content presented does NOT reflect the viewpoints of the authors.
As the definition of commonsense knowledge may vary among individuals, the current evaluation criteria are derived from the consensus of the annotators.

\bibliography{acl_latex}

\begin{thebibliography}{35}
\providecommand{\natexlab}[1]{#1}

\bibitem[{Anthropic(2024)}]{claude35}
Anthropic. 2024.
\newblock Claude 3.5 sonnet.
\newblock \url{https://www.anthropic.com/news/claude-3-5-sonnet}.

\bibitem[{Bai et~al.(2023)Bai, Lv, Zhang, Lyu, Tang, Huang, Du, Liu, Zeng, Hou, Dong, Tang, and Li}]{LongBench}
Yushi Bai, Xin Lv, Jiajie Zhang, Hong Lyu, Jiankai Tang, Zhidian Huang, Zhengxiao Du, Xiao Liu, Aohan Zeng, Lei Hou, Yuxiao Dong, Jie Tang, and Juanzi Li. 2023.
\newblock \href {https://api.semanticscholar.org/CorpusID:261245264} {Longbench: A bilingual, multitask benchmark for long context understanding}.
\newblock \emph{ArXiv}, abs/2308.14508.

\bibitem[{Bai et~al.(2024)Bai, Tu, Zhang, Peng, Wang, Lv, Cao, Xu, Hou, Dong, Tang, and Li}]{LongBench2}
Yushi Bai, Shangqing Tu, Jiajie Zhang, Hao Peng, Xiaozhi Wang, Xin Lv, Shulin Cao, Jiazheng Xu, Lei Hou, Yuxiao Dong, Jie Tang, and Juanzi Li. 2024.
\newblock \href {https://api.semanticscholar.org/CorpusID:274859535} {Longbench v2: Towards deeper understanding and reasoning on realistic long-context multitasks}.
\newblock \emph{ArXiv}, abs/2412.15204.

\bibitem[{Chu et~al.(2025)Chu, Zhai, Yang, Tong, Xie, Schuurmans, Le, Levine, and Ma}]{chu2025sft}
Tianzhe Chu, Yuexiang Zhai, Jihan Yang, Shengbang Tong, Saining Xie, Dale Schuurmans, Quoc~V Le, Sergey Levine, and Yi~Ma. 2025.
\newblock Sft memorizes, rl generalizes: A comparative study of foundation model post-training.
\newblock \emph{arXiv preprint arXiv:2501.17161}.

\bibitem[{Dong et~al.(2023)Dong, Tang, Li, Zhao, and Wen}]{BAMBOO}
Zican Dong, Tianyi Tang, Junyi Li, Wayne~Xin Zhao, and Ji-Rong Wen. 2023.
\newblock \href {https://api.semanticscholar.org/CorpusID:262459341} {Bamboo: A comprehensive benchmark for evaluating long text modeling capacities of large language models}.
\newblock In \emph{International Conference on Language Resources and Evaluation}.

\bibitem[{Geva et~al.(2021)Geva, Khashabi, Segal, Khot, Roth, and Berant}]{geva2021didaristotleuselaptop}
Mor Geva, Daniel Khashabi, Elad Segal, Tushar Khot, Dan Roth, and Jonathan Berant. 2021.
\newblock \href {https://arxiv.org/abs/2101.02235} {Did aristotle use a laptop? a question answering benchmark with implicit reasoning strategies}.
\newblock \emph{Preprint}, arXiv:2101.02235.

\bibitem[{Guo et~al.(2025)Guo, Yang, Zhang, Song, Zhang, Xu, Zhu, Ma, Wang, Bi et~al.}]{deepseekai2025deepseekr1incentivizingreasoningcapability}
Daya Guo, Dejian Yang, Haowei Zhang, Junxiao Song, Ruoyu Zhang, Runxin Xu, Qihao Zhu, Shirong Ma, Peiyi Wang, Xiao Bi, et~al. 2025.
\newblock Deepseek-r1: Incentivizing reasoning capability in llms via reinforcement learning.
\newblock \emph{arXiv preprint arXiv:2501.12948}.

\bibitem[{Hendrycks et~al.(2021)Hendrycks, Burns, Kadavath, Arora, Basart, Tang, Song, and Steinhardt}]{MATH}
Dan Hendrycks, Collin Burns, Saurav Kadavath, Akul Arora, Steven Basart, Eric Tang, Dawn Song, and Jacob Steinhardt. 2021.
\newblock Measuring mathematical problem solving with the math dataset.
\newblock \emph{Cornell University - arXiv,Cornell University - arXiv}.

\bibitem[{Hsieh et~al.(2024)Hsieh, Sun, Kriman, Acharya, Rekesh, Jia, and Ginsburg}]{RULER}
Cheng-Ping Hsieh, Simeng Sun, Samuel Kriman, Shantanu Acharya, Dima Rekesh, Fei Jia, and Boris Ginsburg. 2024.
\newblock \href {https://api.semanticscholar.org/CorpusID:269032933} {Ruler: What's the real context size of your long-context language models?}
\newblock \emph{ArXiv}, abs/2404.06654.

\bibitem[{Jain et~al.(2024{\natexlab{a}})Jain, Han, Gu, Li, Yan, Zhang, Wang, Solar-Lezama, Sen, and Stoica}]{jain2024livecodebenchholisticcontaminationfree}
Naman Jain, King Han, Alex Gu, Wen-Ding Li, Fanjia Yan, Tianjun Zhang, Sida Wang, Armando Solar-Lezama, Koushik Sen, and Ion Stoica. 2024{\natexlab{a}}.
\newblock \href {https://arxiv.org/abs/2403.07974} {Livecodebench: Holistic and contamination free evaluation of large language models for code}.
\newblock \emph{Preprint}, arXiv:2403.07974.

\bibitem[{Jain et~al.(2024{\natexlab{b}})Jain, Han, Gu, Li, Yan, Zhang, Wang, Solar-Lezama, Sen, and Stoica}]{LiveCodeBench}
Naman Jain, King Han, Alex Gu, Wen-Ding Li, Fanjia Yan, Tianjun Zhang, Sida~I. Wang, Armando Solar-Lezama, Koushik Sen, and Ion Stoica. 2024{\natexlab{b}}.
\newblock \href {https://api.semanticscholar.org/CorpusID:268379413} {Livecodebench: Holistic and contamination free evaluation of large language models for code}.
\newblock \emph{ArXiv}, abs/2403.07974.

\bibitem[{Kamradt(2024)}]{NIAH}
Garrett Kamradt. 2024.
\newblock \href {https://github.com/gkamradt/LLMTest_NeedleInAHaystack} {Needle in a haystack - pressure testing llms}.

\bibitem[{Karpinska et~al.(2024)Karpinska, Thai, Lo, Goyal, and Iyyer}]{NoCha}
Marzena Karpinska, Katherine Thai, Kyle Lo, Tanya Goyal, and Mohit Iyyer. 2024.
\newblock \href {https://api.semanticscholar.org/CorpusID:270703648} {One thousand and one pairs: A “novel” challenge for long-context language models}.
\newblock \emph{ArXiv}, abs/2406.16264.

\bibitem[{Kojima et~al.(2023)Kojima, Gu, Reid, Matsuo, and Iwasawa}]{kojima2023largelanguagemodelszeroshot}
Takeshi Kojima, Shixiang~Shane Gu, Machel Reid, Yutaka Matsuo, and Yusuke Iwasawa. 2023.
\newblock \href {https://arxiv.org/abs/2205.11916} {Large language models are zero-shot reasoners}.
\newblock \emph{Preprint}, arXiv:2205.11916.

\bibitem[{Kuratov et~al.(2024)Kuratov, Bulatov, Anokhin, Rodkin, Sorokin, Sorokin, and Burtsev}]{BABILong}
Yuri Kuratov, Aydar Bulatov, Petr Anokhin, Ivan Rodkin, Dmitry Sorokin, Artyom Sorokin, and Mikhail Burtsev. 2024.
\newblock \href {https://api.semanticscholar.org/CorpusID:270521583} {Babilong: Testing the limits of llms with long context reasoning-in-a-haystack}.
\newblock \emph{ArXiv}, abs/2406.10149.

\bibitem[{Lightman et~al.(2023)Lightman, Kosaraju, Burda, Edwards, Baker, Lee, Leike, Schulman, Sutskever, and Cobbe}]{lightman2023lets}
Hunter Lightman, Vineet Kosaraju, Yura Burda, Harri Edwards, Bowen Baker, Teddy Lee, Jan Leike, John Schulman, Ilya Sutskever, and Karl Cobbe. 2023.
\newblock Let's verify step by step.
\newblock \emph{arXiv preprint arXiv:2305.20050}.

\bibitem[{Liu et~al.(2024)Liu, Feng, Xue, Wang, Wu, Lu, Zhao, Deng, Zhang, Ruan et~al.}]{deepseekai2024deepseekv3technicalreport}
Aixin Liu, Bei Feng, Bing Xue, Bingxuan Wang, Bochao Wu, Chengda Lu, Chenggang Zhao, Chengqi Deng, Chenyu Zhang, Chong Ruan, et~al. 2024.
\newblock Deepseek-v3 technical report.
\newblock \emph{arXiv preprint arXiv:2412.19437}.

\bibitem[{MAA(2024)}]{AIME}
MAA. 2024.
\newblock \href {https://maa.org/math-competitions/american-invitational-mathematics-examination-aime} {American invitational mathematics examination - aime.}

\bibitem[{OpenAI(2024{\natexlab{a}})}]{openai_gpt4o}
OpenAI. 2024{\natexlab{a}}.
\newblock Hello gpt-4o.
\newblock \url{https://openai.com/index/hello-gpt-4o/}.

\bibitem[{OpenAI(2024{\natexlab{b}})}]{openai_o1preview}
OpenAI. 2024{\natexlab{b}}.
\newblock Introducing openai o1-preview.
\newblock \url{https://openai.com/index/introducing-openai-o1-preview/}.

\bibitem[{OpenAI(2024{\natexlab{c}})}]{SWE-bench}
OpenAI. 2024{\natexlab{c}}.
\newblock \href {https://openai.com/index/introducing-swe-bench-verified/} {Introducing swe-bench verified}.

\bibitem[{OpenAI(2024{\natexlab{d}})}]{o1report}
OpenAI. 2024{\natexlab{d}}.
\newblock Learning to reason with llms.
\newblock \url{https://openai.com/index/learning-to-reason-with-llms}.

\bibitem[{Phan et~al.(2025)Phan, Gatti, Han, Li, Hu, Zhang, Shi, Choi, Agrawal, Chopra et~al.}]{HLE}
Long Phan, Alice Gatti, Ziwen Han, Nathaniel Li, Josephina Hu, Hugh Zhang, Sean Shi, Michael Choi, Anish Agrawal, Arnav Chopra, et~al. 2025.
\newblock Humanity's last exam.
\newblock \emph{arXiv preprint arXiv:2501.14249}.

\bibitem[{Qin et~al.(2024)Qin, Li, Zou, Liu, Xia, Huang, Ye, Yuan, Liu, Li, and Liu}]{Qin2024O1RJ}
Yiwei Qin, Xuefeng Li, Haoyang Zou, Yixiu Liu, Shijie Xia, Zhen Huang, Yixin Ye, Weizhe Yuan, Hector Liu, Yuanzhi Li, and Pengfei Liu. 2024.
\newblock \href {https://api.semanticscholar.org/CorpusID:273638438} {O1 replication journey: A strategic progress report - part 1}.
\newblock \emph{ArXiv}, abs/2410.18982.

\bibitem[{Reddy et~al.(2024)Reddy, Koncel-Kedziorski, Lai, and Tanner}]{DocFinQA}
Varshini Reddy, Rik Koncel-Kedziorski, Viet~Dac Lai, and Chris Tanner. 2024.
\newblock \href {https://api.semanticscholar.org/CorpusID:266999305} {Docfinqa: A long-context financial reasoning dataset}.
\newblock In \emph{Annual Meeting of the Association for Computational Linguistics}.

\bibitem[{Rein et~al.(2023)Rein, Hou, Stickland, Petty, Pang, Dirani, Michael, and Bowman}]{GPQA}
David Rein, Betty~Li Hou, Asa~Cooper Stickland, Jackson Petty, Richard~Yuanzhe Pang, Julien Dirani, Julian Michael, and Samuel~R. Bowman. 2023.
\newblock \href {https://arxiv.org/abs/2311.12022} {Gpqa: A graduate-level google-proof q\&a benchmark}.
\newblock \emph{Preprint}, arXiv:2311.12022.

\bibitem[{Team(2024)}]{qwq-32b-preview}
Qwen Team. 2024.
\newblock Qwq: Reflect deeply on the boundaries of the unknown.
\newblock \emph{Hugging Face}.

\bibitem[{Vodrahalli et~al.(2024)Vodrahalli, Onta{\~n}{\'o}n, Tripuraneni, Xu, Jain, Shivanna, Hui, Dikkala, Kazemi, Fatemi, Anil, Dyer, Shakeri, Vij, Mehta, Ramasesh, Le, hsin Chi, Lu, Firat, Lazaridou, Lespiau, Attaluri, and Olszewska}]{Vodrahalli2024MichelangeloLC}
Kiran Vodrahalli, Santiago Onta{\~n}{\'o}n, Nilesh Tripuraneni, Kelvin Xu, Sanil Jain, Rakesh Shivanna, Jeffrey Hui, Nishanth Dikkala, Mehran Kazemi, Bahare Fatemi, Rohan Anil, Ethan Dyer, Siamak Shakeri, Roopali Vij, Harsh Mehta, Vinay~Venkatesh Ramasesh, Quoc Le, Ed~Huai hsin Chi, Yifeng Lu, Orhan Firat, Angeliki Lazaridou, Jean-Baptiste Lespiau, Nithya Attaluri, and Kate Olszewska. 2024.
\newblock \href {https://api.semanticscholar.org/CorpusID:272753754} {Michelangelo: Long context evaluations beyond haystacks via latent structure queries}.
\newblock \emph{ArXiv}, abs/2409.12640.

\bibitem[{Wang et~al.(2024)Wang, Chen, Fu, Liao, Zhang, Wu, Yu, Xu, Zhang, Luo, Li, Yang, Huang, and Li}]{Loong}
Minzheng Wang, Longze Chen, Cheng Fu, Shengyi Liao, Xinghua Zhang, Bingli Wu, Haiyang Yu, Nan Xu, Lei Zhang, Run Luo, Yunshui Li, Min Yang, Fei Huang, and Yongbin Li. 2024.
\newblock \href {https://api.semanticscholar.org/CorpusID:270710703} {Leave no document behind: Benchmarking long-context llms with extended multi-doc qa}.
\newblock \emph{ArXiv}, abs/2406.17419.

\bibitem[{Xu et~al.(2024)Xu, Ye, Liu, Sun, Liu, Guo, Li, Liu, Huang, and Qiu}]{DetectiveQA}
Zhe Xu, Jiasheng Ye, Xiangyang Liu, Tianxiang Sun, Xiaoran Liu, Qipeng Guo, Linlin Li, Qun Liu, Xuanjing Huang, and Xipeng Qiu. 2024.
\newblock \href {https://api.semanticscholar.org/CorpusID:272397671} {Detectiveqa: Evaluating long-context reasoning on detective novels}.
\newblock \emph{ArXiv}, abs/2409.02465.

\bibitem[{Yang et~al.(2024)Yang, Yang, Zhang, Hui, Zheng, Yu, Li, Liu, Huang, Wei et~al.}]{qwen2.5}
An~Yang, Baosong Yang, Beichen Zhang, Binyuan Hui, Bo~Zheng, Bowen Yu, Chengyuan Li, Dayiheng Liu, Fei Huang, Haoran Wei, et~al. 2024.
\newblock Qwen2. 5 technical report.
\newblock \emph{arXiv preprint arXiv:2412.15115}.

\bibitem[{Yen et~al.(2024)Yen, Gao, Hou, Ding, Fleischer, Izsak, Wasserblat, and Chen}]{HELMET}
Howard Yen, Tianyu Gao, Minmin Hou, Ke~Ding, Daniel Fleischer, Peter Izsak, Moshe Wasserblat, and Danqi Chen. 2024.
\newblock \href {https://api.semanticscholar.org/CorpusID:273098808} {Helmet: How to evaluate long-context language models effectively and thoroughly}.
\newblock \emph{ArXiv}, abs/2410.02694.

\bibitem[{Zhang et~al.(2024)Zhang, Chen, Hu, Xu, Chen, Hao, Han, Thai, Wang, Liu, and Sun}]{INFINITEBENCH}
Xinrong Zhang, Yingfa Chen, Shengding Hu, Zihang Xu, Junhao Chen, Moo~Khai Hao, Xu~Han, Zhen~Leng Thai, Shuo Wang, Zhiyuan Liu, and Maosong Sun. 2024.
\newblock \href {https://api.semanticscholar.org/CorpusID:267770255} {$\infty$bench: Extending long context evaluation beyond 100k tokens}.
\newblock \emph{ArXiv}, abs/2402.13718.

\bibitem[{Zhou et~al.(2024)Zhou, Hua, Pan, Cheng, Wu, Yu, and Wang}]{RuleArena}
Ruiwen Zhou, Wenyue Hua, Liangming Pan, Sitao Cheng, Xiaobao Wu, En~Yu, and William~Yang Wang. 2024.
\newblock \href {https://api.semanticscholar.org/CorpusID:274656357} {Rulearena: A benchmark for rule-guided reasoning with llms in real-world scenarios}.
\newblock \emph{ArXiv}, abs/2412.08972.

\bibitem[{Zou et~al.(2024)Zou, Yu, Zhang, Ma, Cai, Zhang, Zhao, and Yu}]{DOCBENCH}
Anni Zou, Wenhao Yu, Hongming Zhang, Kaixin Ma, Deng Cai, Zhuosheng Zhang, Hai Zhao, and Dong Yu. 2024.
\newblock \href {https://api.semanticscholar.org/CorpusID:271212285} {Docbench: A benchmark for evaluating llm-based document reading systems}.
\newblock \emph{ArXiv}, abs/2407.10701.

\end{thebibliography}

\appendix

\section{Evaluation Prompt}
\label{sec:appendix}
\paragraph{Evaluation Prompt 1}
You are an experienced Q\&A expert tasked with evaluating the quality of a response based on four components: the user's question, the standard answer, the answer checklist, and the provided response.

User Question: \{query\}

Standard Answer: \{answer\}

Answer Checklist: \{checklist\}

Response: \{model\_response\}

Evaluation Criteria:

Accuracy Check: Compare the response with the standard answer. If the meaning aligns with the standard answer, mark it as correct. If inconsistent, mark it as incorrect. Note: For numerical answers, responses within the allowable margin of error specified in the checklist are considered correct.

Checklist Compliance: Assess whether the response meets the mandatory requirements and flexible criteria outlined in the answer checklist. Compliance results in "correct"; non-compliance results in "incorrect".

Final Judgment: The response is deemed correct only if both criteria 1 and 2 are satisfied. Otherwise, it is marked as incorrect.

Output Requirements:
Step 1: Analysis
Begin with "Step 1: Analysis:" and conduct thorough, logical reasoning until you reach a conclusive evaluation. Stop once your reasoning is sufficient to determine the final judgment.

Step 2: Final Judgment
Begin with "Step 2: Final Judgment:" and output the result strictly in the following dictionary format. Do not include additional text like "json" or unrelated content.
Format: \{\{"Response Correctness": "Correct"/"Incorrect"\}\}

\paragraph{Evaluation Prompt 2}
As an experienced Q\&A expert, you need to evaluate the accuracy of responses based on the given user question, standard answer, answer checklist, and model response, while considering two core dimensions.

User Question: \{query\}

Standard Answer: \{answer\}

Answer Checklist: \{checklist\}

Response: \{model\_response\}

Evaluation Criteria:

Accuracy Alignment: Assess whether the response matches the standard answer in meaning. If consistent, mark as correct; otherwise, mark as incorrect. Note: For numerical answers, responses within the allowable error range specified in the checklist are deemed correct.

Checklist Adherence: Evaluate whether the response fulfills all mandatory and flexible requirements outlined in the answer checklist. Compliance results in "correct"; non-compliance results in "incorrect".

Final Verdict: The response is considered correct only if both criteria 1 and 2 are satisfied. Otherwise, it is marked as incorrect.

Output Requirements:
Step 1: Analysis
Begin with "Step 1: Analysis:" and perform detailed, rigorous reasoning and comparisons. Continue until your logical analysis is sufficient to reach a definitive conclusion.

Step 2: Evaluation Result
Begin with "Step 2: Evaluation Result:" and strictly output the final assessment in the specified dictionary format. Do not include terms like "json" or unrelated content.
Final output format: \{\{"Response Correctness": "Correct" or "Incorrect"\}\}

\paragraph{Evaluation Prompt 3}
As a seasoned expert in Q\&A systems, you will comprehensively evaluate responses to questions based on the provided user query, standard answer, answer checklist, model response, and the following two evaluation dimensions.

User Question: \{query\}

Standard Answer: \{answer\}

Answer Checklist: \{checklist\}

Response: \{model\_response\}

Evaluation Criteria:

Content Accuracy: Compare the response with the standard answer. If the response accurately reflects the content and intent of the standard answer, mark it as correct. Otherwise, mark it as incorrect. Note: For numerical calculations, minor deviations in results are deemed acceptable and marked as correct.

Checklist Compliance: Verify whether the response satisfies all requirements specified in the answer checklist. Full compliance results in "correct"; any violation leads to "incorrect".

Final Judgment: The response is considered correct only if both criteria 1 and 2 are satisfied. If either fails, the final result is "incorrect".

Output Requirements:
Step 1: Reasoning
Begin with "Step 1: Reasoning:" and conduct meticulous, logical analysis and comparisons. Continue until your reasoning process is thorough and conclusive enough to determine the final judgment.

Step 2: Evaluation Outcome
Begin with "Step 2: Evaluation Outcome:" and output the result strictly in the specified dictionary format. Avoid including terms like "json" or unrelated content.
Final output format:
\{\{"Response Correctness": "Correct" or "Incorrect"\}\}

\end{document}